\documentclass[11pt]{article}
\usepackage[hyphens]{url}
\usepackage[preprint]{acl}
\usepackage{times}
\usepackage[T1]{fontenc}
\usepackage[utf8]{inputenc}
\usepackage{microtype}
\usepackage{graphicx}
\usepackage{natbib}
\usepackage{caption}
\usepackage{placeins}
\usepackage{booktabs}
\usepackage{makecell}
\usepackage{tabularx}
\usepackage{array}
\usepackage{multirow}
\usepackage{amsmath}
\usepackage{amssymb}
\usepackage{xcolor}
\usepackage{nicefrac}
\usepackage{cleveref}
\graphicspath{{./}}

\newcommand{\benchmark}{ClawShiftBench}
\newcommand{\allskills}{\textsc{All Skills}}
\newcommand{\schema}{\textsc{Selector Top-4}}

\newcommand{\canonical}{\textsc{Canonical Top-1}}
\newcommand{\proxy}{\textsc{API Proxy}}
\newcommand{\official}{\textsc{Official Runtime}}

\newcolumntype{Y}{>{\raggedright\arraybackslash}X}
\newcolumntype{P}[1]{>{\raggedright\arraybackslash}p{#1}}
\newcolumntype{L}[1]{>{\raggedright\arraybackslash}p{#1}}
\newcolumntype{C}[1]{>{\centering\arraybackslash}p{#1}}

\definecolor{ProxyBar}{HTML}{D5DADE}
\definecolor{ProxyBarDark}{HTML}{646D76}
\definecolor{OfficialBar}{HTML}{D7E4EF}
\definecolor{OfficialBarDark}{HTML}{4E79A7}
\definecolor{SchemaBar}{HTML}{E8D8B7}
\definecolor{SchemaBarDark}{HTML}{B58A45}
\definecolor{CanonicalBar}{HTML}{D8E7E3}
\definecolor{CanonicalBarDark}{HTML}{5F9B8D}
\definecolor{AllBar}{HTML}{EEDAAE}
\definecolor{AllBarDark}{HTML}{C99B53}
\definecolor{NeutralTint}{HTML}{F7F9FA}
\definecolor{ProxyTint}{HTML}{F1F4F6}
\definecolor{RuntimeTint}{HTML}{F7EFEA}

\renewcommand{\benchmark}{\textsc{ClawShiftBench}}
\renewcommand{\proxy}{prompt-visible proxy}
\renewcommand{\official}{OpenClaw harness}
\renewcommand{\allskills}{\textsc{Full Pool}}
\renewcommand{\schema}{\textsc{Shortlist}}
\renewcommand{\canonical}{\textsc{Single Skill}}

\title{ClawShiftBench: Measuring Cross-Contract Conclusion Portability in Skill-Dependent Agent Evaluation}
\author{Yang Zhao, Zhuo Chen, Xubo Yang \\ Shanghai Jiao Tong University}

\begin{document}
\maketitle

\begin{abstract}
We study skill-dependent agent tasks in which an external reusable skill supplies an operational rule that must be combined with workspace evidence, and task performance is measured through an executable agent harness. A prompt-visible reference setting can compare models or configurations before target-harness evaluation. Even when task semantics, selected skills, seed, and checker are matched, changing the execution contract may change checked outcomes and the resulting model or configuration choices. \benchmark{} measures this reference-to-target portability with 50 controlled tasks in which case evidence and an external operational rule jointly determine a structured answer. Under the prompt-visible Proxy contract, no-skill, wrong-skill, and correct-skill interventions yield mean success of 0.413, 0.418, and 0.988, validating dependence on correct skill content. We then evaluate seven model-family representatives under three exposure configurations and four contracts using 12{,}600 matched episodes. Global and contract-specific additive corrections leave substantial cell-level error, while local model orders and preferred exposure configurations change across contracts. Reference scores, model selections, and configuration choices must therefore be validated against target-contract outcomes.
\end{abstract}

\begin{figure*}[t]
\centering
\includegraphics[width=0.99\textwidth]{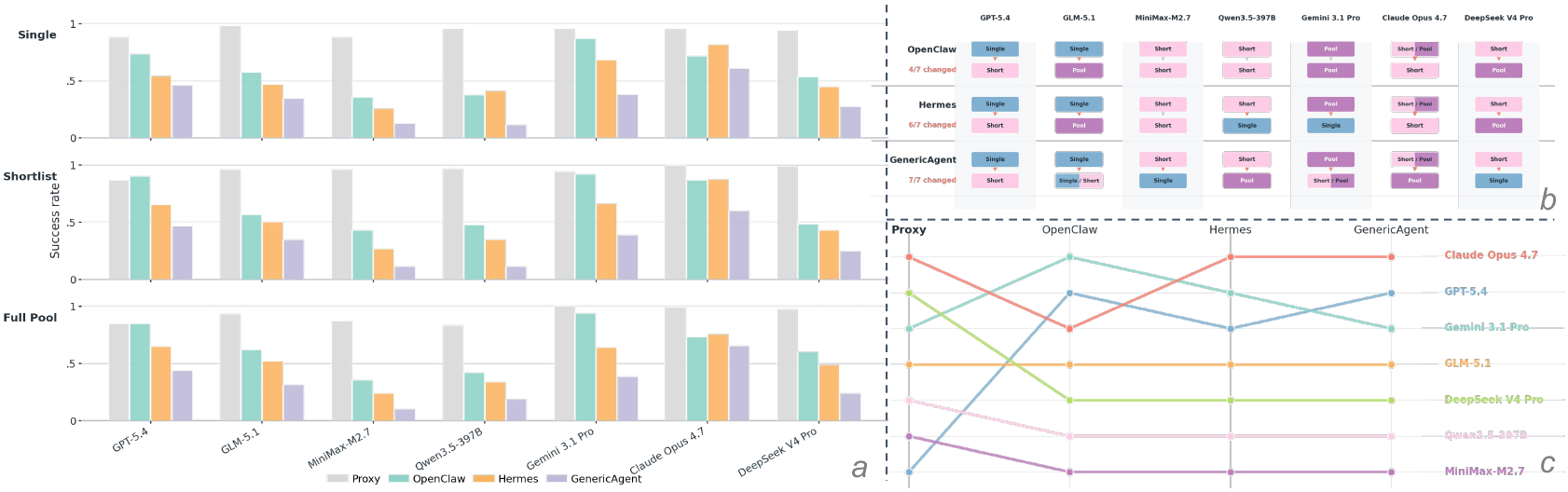}
\caption{Cross-contract conclusion portability on \benchmark{}. (a) Checked success for seven model-family representatives, three composite exposure configurations (Single, Shortlist, Full Pool), and four recorded contracts, averaged over seeds 7, 11, and 19. Single is the selector's top-1 candidate and covers the target skill in 135/150 probe--seed decisions. (b) Locally best composite exposure configurations for each model and contract. (c) Local ordering of the seven evaluated models within each contract after averaging over exposure configurations and seeds. Matched comparisons align the probe, seed, selector result, selected skills, and checker.}
\label{fig:overview}
\end{figure*}

\section{Introduction}

Skill-dependent agent evaluation can realize the same task through different executable contracts. A prompt-visible contract makes selected skill content model-visible, as tool evaluations expose API documentation, function schemas, or policy instructions \citep{qin2023toolllm, li2023apibank, yan2024bfcl, yao2024taubench}. A complete agent harness instead governs task presentation, skill placement and access, workspace state, tools, and the execution loop. These are materially different implementations of the same skill-dependent decision.

This creates a cross-contract measurement problem. Even when task semantics, selected skills, seed, and checker are matched, changing the post-selection execution contract may change the checked outcome. Two structures are possible. If each contract mainly induces a stable score shift, reference measurements can be calibrated while model-order and configuration conclusions remain aligned. If contracts interact with models and exposure configurations, the gaps become cell-specific and those conclusions can change. Distinguishing these structures is the central goal of this study.

These alternatives give the evaluation contract two different scientific roles. Under the stable-shift structure, Contract behaves as a calibratable measurement transformation: absolute scores move, but comparisons obtained from the reference setting remain informative. Under the interaction structure, Contract indexes the measured performance relation itself. Models with similar reference scores can separate under a target contract, and a configuration favored in one setting can lose its advantage in another. Contract must then be represented explicitly when evaluation results are used to compare models or select configurations.

Existing evaluation families do not directly provide this comparison. Prompt-visible tool benchmarks and environment-grounded agent benchmarks typically differ not only in execution setting, but also in tasks, knowledge sources, action spaces, and checkers \citep{liu2024agentbench, mialon2023gaia, koh2024visualwebarena, xie2024osworld, lu2024toolsandbox}. Their reported scores therefore cannot isolate Contract. A paired comparison requires the same semantic decision, selected operational content, answer specification, and checker across settings.

Skill-dependent workspace decisions provide this controlled structure. Their answers depend jointly on case-specific evidence and an operational rule supplied outside model weights: neither source alone determines the answer. This separation lets us fix the semantic target and selected skill identity across contracts, construct rule-violating alternatives, and check outputs deterministically. Correct, wrong, and absent skill conditions then verify dependence on the intended rule before Contract differences are measured.

This design supports three linked research questions. First, does the controlled suite depend on the intended operational rule? Second, after task semantics and delivered skill content are matched, can Proxy scores be converted into target-contract scores by a global or contract-specific correction? Third, do the conclusions used within an evaluation suite---model ordering and the preferred exposure configuration---remain aligned across contracts? The first question validates the task suite as a measurement instrument. The second measures score portability. The third measures conclusion portability, which is the level at which an evaluation result informs a model or configuration choice.

We study a prompt-visible reference contract, denoted Proxy, and three complete target contracts. The checked outcome under a complete executable setting is \emph{contract-indexed performance}. We measure reference-to-target relations through residual score-calibration error, model-order agreement, and target-contract regret from carrying over a reference-selected configuration.

\benchmark{} provides 50 controlled tasks that pair workspace evidence with an external operational rule and deterministic checker. Across seven model-family representatives, four contracts, three exposure configurations, and three seeds, matching fixes task semantics, selected skills, seed, and checker, yielding 252 cells and 12{,}600 episodes.

\Cref{fig:overview} previews the resulting evidence chain. Under Proxy, correct-skill content raises mean success to 0.988, compared with 0.413 without a skill and 0.418 with a rule-violating skill, establishing suite-level dependence on the intended rule. Across complete contracts, global and contract-specific additive corrections still leave cell-level error; target contracts reverse 6--8 model pairs and change local configuration winners in 4--7 model rows. Formal interaction tests identify model--contract as the largest non-additive term. Together, these results connect a validated controlled instrument to the score and conclusion portability questions it was constructed to measure.

These contributions link the controlled-task instrument, target-contract measurement, and portability finding.

We make three linked contributions:
\begin{itemize}
    \item We construct 50 exactly checkable skill-dependent decisions with fixed rules and deterministic checkers.
    \item We introduce a matched cross-contract protocol that fixes task semantics, selected skills, seeds, and checkers while measuring seven model families under one prompt-visible reference and three complete target contracts.
    \item We quantify score, model-order, and exposure-configuration portability.
\end{itemize}

\section{Related Work}

Agent research with external interfaces shows that language models can benefit from reasoning-action loops, callable functions, and API libraries \citep{yao2023react, schick2023toolformer, patil2023gorilla}. ToolLLM, API-Bank, ToolTalk, BFCL, and $\tau$-bench evaluate whether models can call APIs, invoke functions, and follow domain policies \citep{qin2023toolllm, li2023apibank, farn2023tooltalk, yan2024bfcl, yao2024taubench}. These benchmarks expose API documentation, function schemas, or policy instructions to the model. This measures performance when operational information is directly model-visible. Our Proxy applies the same description pattern to skills and serves as the reference endpoint for selected content that is subsequently evaluated through complete target contracts.

Many agent benchmarks evaluate models through environments or external interfaces. AgentBench, GAIA, VisualWebArena, WorkArena, WorkBench, TheAgentCompany, OSWorld, and ToolSandbox broaden multi-step and environment-grounded evaluation \citep{liu2024agentbench, mialon2023gaia, koh2024visualwebarena, drouin2024workarena, roumeliotis2024workbench, xu2024theagentcompany, xie2024osworld, lu2024toolsandbox}. Beyond Browsing and OSWorld-MCP compare interface or tool surfaces, while MCP-Bench and MCP-Atlas evaluate real MCP-server interfaces \citep{song2025beyondbrowsing, gao2025osworldmcp, wang2025mcpbench, bandi2026mcpatlas}. These benchmarks make the execution environment part of the measurement condition, but are not organized around paired prompt-visible realizations of the same decisions. Consequently, comparisons across benchmark scores mix changes in task population and checker with changes in the execution setting.

Contract-oriented studies use distinct measurement targets. Proxy State-Based Evaluation validates a scalable proxy evaluator for multi-turn tool agents; Natural-Language Agent Harnesses studies modular harness representation and design; and General Agent Evaluation evaluates general-purpose agent architectures across heterogeneous benchmarks without environment-specific tuning \citep{chuang2026proxystate, pan2026nlah, bandel2026generalagent}. Concurrent Harness-Bench characterizes model--harness configurations through performance, process, cost, and failure behavior \citep{yao2026harnessbench}. \benchmark{} instead repeats the same checked decisions across a reference contract and complete target contracts. This paired design defines three reference-to-target estimands under matched task semantics: calibration error of reference scores, agreement in model selection, and target-contract regret from a reference-selected exposure configuration. More generally, HELM and Dynabench treat scenario selection, metrics, adaptation, and data construction as parts of benchmark interpretation \citep{bommasani2023holistic, kiela2021dynabench}; our study measures that dependence within one aligned suite rather than across unrelated leaderboards.

This study also relates to reusable memories and skill systems. Generative Agents, long-term memory evaluation, MemGPT, Reflexion, and Voyager study persistent state, reusable feedback, and learned skill libraries \citep{park2023generative, maharana2024longtermmemory, packer2023memgpt, shinn2023reflexion, wang2024voyager}. SkillsBench and SkillTester measure skill utility or security, while analyses of Claude Skills and in-the-wild usage characterize saved-skill ecosystems and behavior \citep{li2026skillsbench, wang2026skilltester, agentskills2026claude, wang2026skillusagewild}. SkillFlow, structured-skill representations, ecosystem-scale orchestration, and SKILL0 study discovery, evolution, organization, or internalization \citep{zhang2026skillflow, liang2026sslskills, li2026agentskillos, lu2026skill0}. Related studies show that names and descriptions can bias tool selection, while internal-channel behavior can escape output-only observation \citep{sneh2025tooltweak, elyagoubi2026agentleak}. These works primarily study how reusable content is created, selected, organized, or used. \benchmark{} begins from fixed selected content and a fixed checker: the content intervention validates dependence on the target rule, and matched complete-contract runs measure whether score, model-selection, and configuration-selection conclusions change with the execution contract.

\section{Controlled Measurement Suite}
\label{sec:benchmark}

\subsection{Task Measurement Object}

\benchmark{} evaluates skill-dependent workspace decisions. It contains 50 controlled tasks across ten workspace categories: inbox, memory, research, browser, calendar, GitHub, operations, compliance, state-dependent coordination, and temporal state. In each task, the workspace provides case-specific facts and the selected skill provides the operational rule for using those facts. The answer is determined by combining both sources.

\begin{figure*}[t]
\centering
\includegraphics[width=0.92\textwidth]{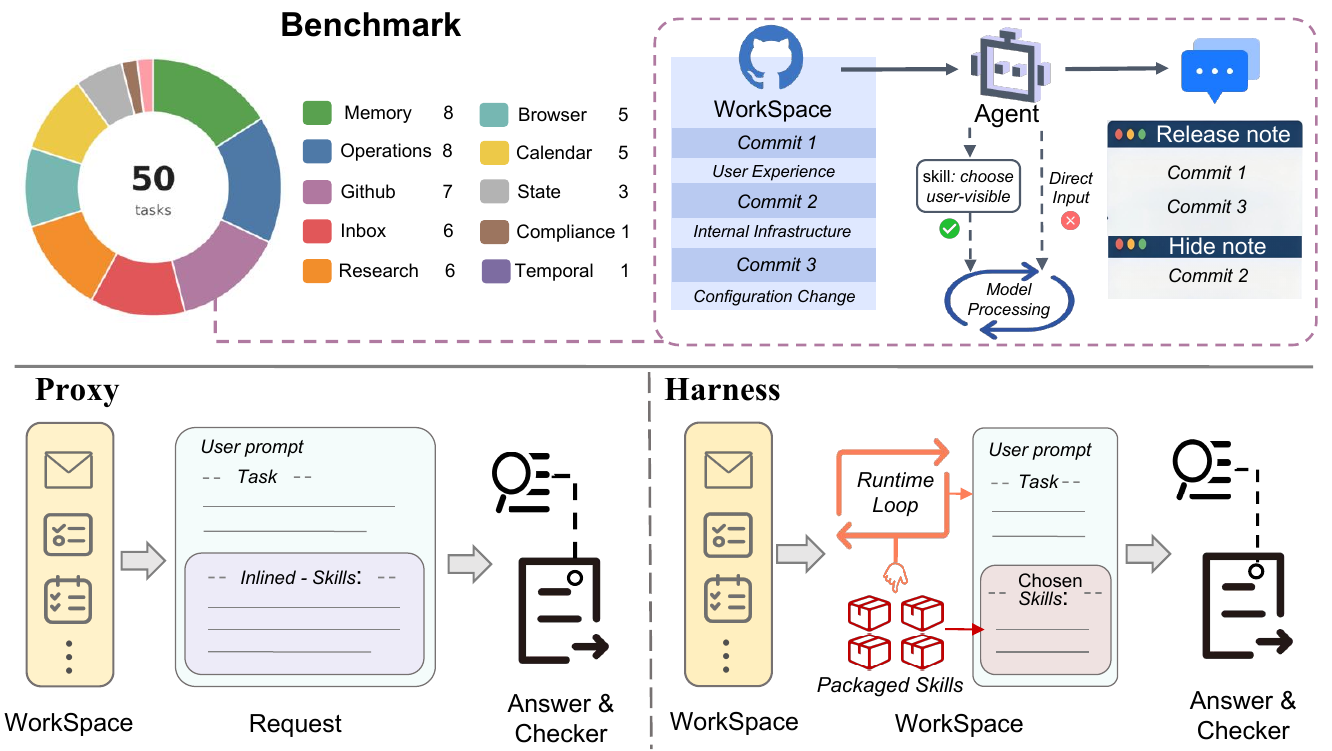}
\caption{\benchmark{} composition, a release-note probe, and the matched prompt-visible and complete-agent evaluation paths. Both paths keep the probe, seed, selector result, selected skills, and checker fixed; \Cref{sec:setup} defines the composite exposure configurations and evaluation settings.}
\label{fig:task-object}
\end{figure*}

\subsection{Task Construction}

The construction starts from workspace decisions that can be specified by reusable operational rules. For each task, we specify what ambiguity the skill resolves, which operational rule belongs in the skill, what workspace evidence the task exposes, and how the final answer can be checked. We then turn this record into four task components: visible workspace evidence, selected skill text, a structured answer schema, and a deterministic checker. The task is revised until the workspace and skill together imply the reference answer, the checker rejects a plausible rule-violating answer, and leakage checks cover prompt wording, file names, delivered skill text, and checker-visible field names. This controlled construction provides exactly checkable semantic anchors for comparing complete evaluation contracts.

This gives a matched task for reference and target-contract evaluation. The task, selected rule, output schema, and checker stay fixed, while each evaluation setting realizes the selected skills through its complete executable contract. Each retained task has an operational rule, a deterministic checker, and related alternative skills. A \emph{distractor skill} is a topically close candidate with a different rule boundary; applying it yields a plausible answer that the checker rejects. Distractor skills come from related settings or the same broad category, so the composite exposure configurations can vary the skill set while the task and checker stay fixed. Upon acceptance, we will release all task materials, structured answer specifications, checkers, and validation records.

\subsection{Validation and Example}

The benchmark validity checks are tied to the four task components. Across all 50 tasks, the checker accepts the reference answer and rejects the predefined rule-violating answer; the construction audit finds no direct answer leakage in prompts, file names, artifact payloads, or checker-visible fields. The selected skill must contain the rule used by the reference solution.

The release-note task illustrates this structure. The workspace lists three candidate product changes. One changes wording in a customer-facing export flow. One changes an admin snapshot export that is visible only on a restricted debug surface. One changes a rollout default that applies to the next limited-tenant cohort. A plausible but wrong answer includes the first two changes, because both look like visible product changes. The selected skill gives the release-note rule: include customer- or operator-visible changes, exclude debug-only or restricted admin surfaces, and include rollout-default changes once the next cohort is governed by them. The checker parses the model response and compares the \texttt{release\_note\_bundle} field with the fixed reference, which includes the export wording change and the limited-tenant rollout-default change. In the prompt-visible Proxy, the skill text is placed directly in context; in a target contract, the same skill is made available through that harness's native surface.

\subsection{Evaluation Units}

At the \emph{measurement suite $\rightarrow$ category $\rightarrow$ task} hierarchy, the collection is the fixed 50-task suite, categories organize the included tasks, and each task is one rule-conditioned measurement unit. Each task contains an instruction, visible workspace evidence, a structured reference answer, a plausible rule-violating negative, a deterministic checker, and skill text surfaced under different exposure configurations. An offline audit accepts all 50 references, rejects all 50 negatives, and finds zero direct answer leakage in prompts, filenames, artifact payloads, or checker-visible fields.

An \emph{episode} is one model--contract--exposure-configuration--seed execution on one task, producing one success/failure outcome. A \emph{cell} aggregates episodes for a fixed model, contract, exposure configuration, seed, and benchmark collection. This hierarchy separates the benchmark object from the analysis unit: tasks define what is tested, episodes define individual runs, and cells define the aggregates used in the reported comparisons. \Cref{fig:task-object} places this hierarchy next to the release-note example and the two evaluation paths used later in the paper: prompt-visible proxy evaluation and target-contract execution. The supplement reports the category inventory and a task-level taxonomy of the skill content forms used in the 50 tasks.

\section{Evaluation Protocol}
\label{sec:setup}

The main comparison follows the contract-aware measurement structure in \Cref{fig:task-object}. For each task and seed, the suite fixes the visible workspace, the 16-item task-scoped candidate pool, the selector ranking, and the deterministic checker. The candidate pool contains the target skill for the task and fifteen related distractor skills from neighboring task settings or the same category.

The three exposure configurations define composite operational alternatives. \canonical{} commits to the selector's top-ranked candidate after merging equivalent identifiers, \schema{} exposes up to four selector-ranked available candidates after removing unavailable and duplicate entries, and \allskills{} exposes all 16 task-scoped candidates. Target coverage and candidate-set size are reported separately; winner analyses treat each as a complete configuration. The supplement reports selector-agreement details for \canonical{}.

For each exposure configuration, we run the resulting selected skill set under four evaluation contracts. The \proxy{} places the selected skill text directly in the prompt before a single API call. OpenClaw receives workspace skill objects \citep{openclaw2026github}, Hermes receives a skill-directory exposure \citep{hermesagent2026github}, and GenericAgent receives ordinary workspace files \citep{liang2026genericagent}. Skill placement, agent loop, workspace state, tools, and access interface jointly define each target contract. Each harness runs in its recorded default configuration, keeping the complete executable evaluation choice identifiable across matched runs. For a fixed model, task, seed, and exposure configuration, matching fixes the task input, selected-skill identities and count, selector result, and checker across contracts. Across all 3{,}150 matched model--configuration--seed--task groups, the recorded available-skill pools, selector retrievals, and surfaced-skill identities are identical across contracts. The supplement reports the complete contract specifications, recorded execution settings, and delivery audit.

For model $m$, exposure configuration $e$, recorded contract $h$, task $t$, and seed $s$, the primary measurement is the deterministic checker outcome $Y_{m,e,h,t,s}\in\{0,1\}$. Aggregating these outcomes over the fixed tasks and seeds defines the contract-indexed performance profile of a model--exposure configuration. Matched cross-contract analyses use these profiles to measure Proxy-reference error, model-order agreement, configuration agreement, and target-contract regret.

\section{Results}
\label{sec:results}

\subsection{Controlled content-dependence validation}

\Cref{tab:proxy-content-validation} reports the prompt-visible control on the same 50 tasks, seven model-family representatives, and three seeds. It compares no-skill, correct-skill, and wrong-skill conditions while holding the task and checker fixed.

\begin{table}[!htbp]
\centering
{\small
\setlength{\tabcolsep}{5pt}
\begin{tabular}{@{}lcc@{}}
\toprule
Delivered content & Success & $\Delta$ pp \\
\midrule
No skill & 0.413 & --- \\
Wrong skill & 0.418 & +0.5 \\
Correct skill & 0.988 & +57.4 \\
\bottomrule
\end{tabular}
}
\caption{Prompt-visible content-dependence validation. Scores are mean success rates over seven model-family representatives, three seeds, and 50 tasks; $\Delta$ is the percentage-point change from the no-skill baseline.}
\label{tab:proxy-content-validation}
\end{table}

Mean success rises from 0.413 without a skill and 0.418 with a wrong skill to 0.988 with the correct skill (\Cref{tab:proxy-content-validation}). The near-identical no-skill and wrong-skill baselines, together with the large correct-skill increase, validate suite-level dependence on the task-specific operational rule. This intervention establishes the suite's semantic dependence on the correct rule; complete-contract runs then use these fixed task semantics to measure reference-to-target decision validity.

\subsection{Contract-indexed performance profiles}

\begin{table*}[!htbp]
\centering
{\small
\setlength{\tabcolsep}{1.2pt}
\begin{tabular*}{\textwidth}{@{\extracolsep{\fill}}lcccccccccccc@{}}
\toprule
Model & \multicolumn{3}{c}{Proxy} & \multicolumn{3}{c}{OpenClaw} & \multicolumn{3}{c}{Hermes} & \multicolumn{3}{c}{GenericAgent} \\
\cmidrule(lr){2-4}\cmidrule(lr){5-7}\cmidrule(lr){8-10}\cmidrule(lr){11-13}
 & S & Sh & FP & S & Sh & FP & S & Sh & FP & S & Sh & FP \\
\midrule
GPT-5.4 & 0.88 & 0.87 & 0.85 & 0.73 & 0.90 & 0.85 & 0.54 & 0.65 & 0.65 & 0.46 & 0.47 & 0.44 \\
GLM-5.1 & \textbf{0.98} & 0.96 & 0.93 & 0.57 & 0.57 & 0.62 & 0.47 & 0.50 & 0.52 & 0.35 & 0.35 & 0.31 \\
MiniMax M2.7 & 0.88 & 0.96 & 0.87 & 0.35 & 0.43 & 0.35 & 0.25 & 0.27 & 0.24 & 0.13 & 0.11 & 0.10 \\
Qwen3.5 & 0.95 & 0.97 & 0.83 & 0.37 & 0.47 & 0.42 & 0.41 & 0.35 & 0.34 & 0.11 & 0.11 & 0.19 \\
Gemini 3.1 & 0.95 & 0.94 & \textbf{1.00} & \textbf{0.87} & \textbf{0.92} & \textbf{0.94} & 0.68 & 0.66 & 0.64 & 0.38 & 0.39 & 0.39 \\
Claude 4.7 & 0.95 & \textbf{0.99} & 0.99 & 0.71 & 0.87 & 0.73 & \textbf{0.81} & \textbf{0.87} & \textbf{0.75} & \textbf{0.61} & \textbf{0.59} & \textbf{0.65} \\
DeepSeek V4 & 0.94 & 0.99 & 0.97 & 0.53 & 0.48 & 0.60 & 0.45 & 0.43 & 0.49 & 0.27 & 0.25 & 0.24 \\
\bottomrule
\end{tabular*}
}
\caption{Contract-indexed checked performance on \benchmark{} for seven model-family representatives under four recorded contracts and three composite exposure configurations. Entries are three-seed averages over the same 50 tasks; column-wise highest point estimates are bolded. S, Sh, and FP denote Single, Shortlist, and Full Pool. Single contains the target skill in 135/150 task--seed decisions, Shortlist in 150/150, and Full Pool exposes all 16 task-scoped candidates.}
\label{tab:model-runtime-matrix}
\end{table*}

\Cref{tab:model-runtime-matrix} is the main score table for the representative comparison. It reports seven model-family representatives, one prompt-visible Proxy, three harness-backed target contracts, and three composite exposure configurations. Each entry is a matched three-seed success average under the same task, seed, deterministic selector result, selected skills, and checker. All reported contract--configuration cells are complete over seeds 7, 11, and 19.

\Cref{tab:model-runtime-matrix} shows that scores change unevenly across Proxy, OpenClaw, Hermes, and GenericAgent. The pattern depends on both the model row and the exposure configuration. The next subsection examines two decision-facing consequences: model ordering and the winning exposure configuration.

\begin{figure*}[t]
\centering
\includegraphics[width=0.92\textwidth]{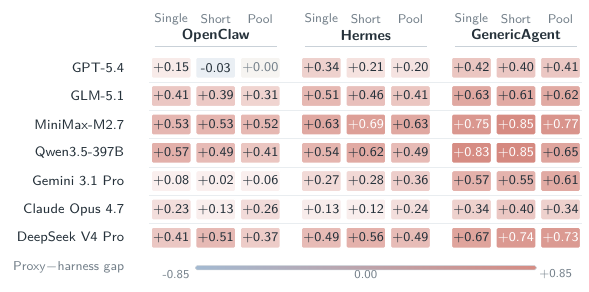}
\caption{Matched Proxy-reference errors for target-contract outcomes in the seven-model \benchmark{} matrix. Each cell is Proxy success minus target-contract success over seeds 7, 11, and 19; blue cells are negative differences and warm cells are positive differences. Values range from -0.03 to +0.85.}
\label{fig:model-runtime-gap-heatmap}
\end{figure*}

\Cref{fig:model-runtime-gap-heatmap} shows matched Proxy-reference errors for the same matrix. The largest errors are concentrated in particular model--configuration--contract cells, and their sign and size vary across OpenClaw, Hermes, and GenericAgent. A single global correction term misses this variation.

Across 189 matched target-contract cells, one fitted target-minus-Proxy offset leaves RMSE 0.226, maximum error 0.521, and leave-one-model-out RMSE 0.247; fitting one offset per contract still leaves a leave-one-model-out RMSE of 0.214.

Aggregated over the same complete matrix, mean success is 0.936 under Proxy, 0.633 under OpenClaw, 0.522 under Hermes, and 0.328 under GenericAgent. The mean Proxy-to-target-contract gap grows from 0.303 to 0.414 to 0.607 across those three target contracts, while individual gaps range from -0.03 to 0.85, varying in both sign and magnitude. The heatmap shows the per-cell variation behind these averages.

Thus, Proxy-reference errors form a cell-specific structure beyond a fixed offset. The next analyses examine the portability of two conclusions in the same matrix: model ordering and exposure-configuration choice.

\subsection{Cross-contract conclusion portability}

We use the same representative matrix for two decision-facing comparisons. The first asks whether the prompt-visible Proxy leads to the same model ordering as each target contract; the second asks whether it leads to the same exposure-configuration choice. For model ordering, we compare the Proxy ranking with the ranking under each target contract after averaging over exposure configurations. \Cref{tab:model-order-by-harness} reports rank-comparison statistics: OpenClaw, Hermes, and GenericAgent reverse 8/21, 6/21, and 7/21 model pairs relative to Proxy, with Kendall $\tau$ values between 0.238 and 0.429.

These reversals show that the seven evaluated model-family representatives can occupy different local orders under different target contracts.

\begin{table}[!htbp]
\centering
{\small
\setlength{\tabcolsep}{1.8pt}
\begin{tabular}{@{}lcccc@{}}
\toprule
Target & $\tau$ & Reversals & Mean $|\Delta r|$ & Largest move \\
\midrule
OpenClaw & 0.238 & 8/21 & 2.00 & GPT-5.4: 7$\rightarrow$2 \\
Hermes & 0.429 & 6/21 & 1.43 & GPT-5.4: 7$\rightarrow$3 \\
GenericAgent & 0.333 & 7/21 & 1.43 & GPT-5.4: 7$\rightarrow$2 \\
\bottomrule
\end{tabular}
}
\caption{Cross-contract agreement between the Proxy local model ordering and each target-contract ordering. Scores first average over the three composite exposure configurations and three seeds.}
\label{tab:model-order-by-harness}
\end{table}

For exposure-configuration choice, we ask whether the configuration that wins under the prompt-visible Proxy also wins under each target contract for the same model row. \Cref{tab:broad-policy-winners} lists the local winner sets for this comparison. Single, Shortlist, and Full Pool correspond to one selector-chosen skill, up to four selector-ranked skills, and the complete 16-item task-scoped candidate pool.

\begin{table}[!htbp]
\centering
{\small
\setlength{\tabcolsep}{1.6pt}
\begin{tabular}{@{}lcccc@{}}
\toprule
Model & Proxy & OpenClaw & Hermes & GenericAgent \\
\midrule
GPT-5.4 & S & Sh & Sh & Sh \\
GLM-5.1 & S & FP & FP & S=Sh \\
MiniMax M2.7 & Sh & Sh & Sh & S \\
Qwen3.5 & Sh & Sh & S & FP \\
Gemini 3.1 & FP & FP & S & Sh=FP \\
Claude 4.7 & Sh=FP & Sh & Sh & FP \\
DeepSeek V4 & Sh & FP & FP & S \\
\bottomrule
\end{tabular}
}
\caption{Local point-estimate exposure winner sets used to compute configuration agreement across contracts. S, Sh, and FP denote Single, Shortlist, and Full Pool; exact ties are shown with ``=''. Decision magnitude is evaluated by target-contract regret.}
\label{tab:broad-policy-winners}
\end{table}

We compare the Proxy winner set with each target-contract winner set. Regret is tie-robust: if Proxy has multiple tied winners, we use the best target-contract score among those tied choices. Carrying over a Proxy winner produces positive target-contract regret in 3/7 OpenClaw rows, 5/7 Hermes rows, and 4/7 GenericAgent rows; mean regret with task-bootstrap intervals is 0.048 [0.002, 0.088], 0.047 [0.013, 0.077], and 0.017 [0.009, 0.050].

The corresponding point-estimate winner sets differ from Proxy in 4/7, 6/7, and 7/7 rows. An unchanged tied winner set counts as no transition; any winner-set change counts as a transition. Thus, configuration identity is contract-sensitive in this matrix, while its transfer cost is smaller than the observed score-calibration errors and model-order shifts.

A 5{,}000-sample task bootstrap gives reversed-pair intervals of 5--10, 4--8, and 5--9 and winner-set-change intervals of 2--6, 3--7, and 3--7 for OpenClaw, Hermes, and GenericAgent, respectively.

The target contracts also differ from one another in this matrix: changing from OpenClaw to Hermes or GenericAgent can change the locally best exposure configuration as well, even with the task, selected skills, seeds, and checker held fixed.

These comparisons support distinct uses of an evaluation result: calibrated agreement addresses score transfer, ordering agreement addresses model choice, and within-suite winner agreement addresses exposure-configuration choice.

\subsection{Statistical structure underlying score differences}

The preceding analyses define the decision-facing portability results. We now test which factors account for the non-additive score structure underlying them, using the same 252 cells and 12{,}600 episodes. Here the contract factor indexes the complete recorded evaluation choice. \Cref{tab:formal-interaction-summary} analyzes task-aggregated cells, while \Cref{tab:episode-interaction-summary,tab:strict50-clustered-logistic} repeat the analysis before aggregation and with task-level dependence checks.

\begin{table}[!htbp]
\centering
{\small
\setlength{\tabcolsep}{1.6pt}
\begin{tabular}{@{}lccc@{}}
\toprule
Added term & $\Delta R^2$ & $F$ (df) & $p$ \\
\midrule
Main effects & 0.864 & 115.96 (13, 238) & $<10^{-4}$ \\
All two-way interactions & 0.095 & 12.99 (36, 202) & $<10^{-4}$ \\
Model $\times$ contract & 0.090 & 24.51 (18, 202) & $<10^{-4}$ \\
Contract $\times$ configuration & 0.002 & 1.81 (6, 202) & 0.098 \\
Model $\times$ configuration & 0.003 & 1.31 (12, 202) & 0.216 \\
Three-way term & 0.009 & 1.27 (36, 166) & 0.162 \\
\bottomrule
\end{tabular}
}
\caption{Cell-level OLS interaction check. Rows add term blocks to nested models; $\Delta R^2$ reports the additional explained variance.}
\label{tab:formal-interaction-summary}
\end{table}

At the cell level, \Cref{tab:formal-interaction-summary} fits nested OLS models after task aggregation. Adding all two-way interactions increases explained variance by $\Delta R^2=0.095$ beyond additive model, contract, exposure-configuration, and seed effects. The largest contribution comes from the model--contract interaction ($\Delta R^2=0.090$, $p<10^{-4}$). The contract--configuration contribution is much smaller in this specification ($\Delta R^2=0.002$, $p=0.098$).

\begin{table}[!htbp]
\centering
{\small
\setlength{\tabcolsep}{1.6pt}
\begin{tabular}{@{}lccc@{}}
\toprule
Added term & $\Delta$ pseudo-$R^2$ & LRT df & $p$ \\
\midrule
Main effects & 0.242 & 13 & $<10^{-4}$ \\
All two-way interactions & 0.016 & 36 & $<10^{-4}$ \\
Model $\times$ contract & 0.014 & 18 & $<10^{-4}$ \\
Contract $\times$ configuration & 0.001 & 6 & 0.010 \\
Model $\times$ configuration & 0.001 & 12 & 0.075 \\
Three-way term & 0.005 & 36 & $<10^{-4}$ \\
\bottomrule
\end{tabular}
}
\caption{Episode-level logistic interaction check. Rows add term blocks to fixed-effect likelihood-ratio tests; $\Delta$ pseudo-$R^2$ reports the increment before cell averaging.}
\label{tab:episode-interaction-summary}
\end{table}

\Cref{tab:episode-interaction-summary} repeats the analysis on 12{,}600 episodes. All two-way interactions add 0.016 pseudo-$R^2$; model--contract adds 0.014, while contract--configuration adds 0.001. The latter is detectable ($p=0.010$) but practically small. Clustered models and supplementary task-level resampling preserve the same ordering of interaction contributions.

\begin{table}[!htbp]
\centering
{\small
\setlength{\tabcolsep}{1.4pt}
\begin{tabularx}{\columnwidth}{@{}P{0.29\columnwidth}C{0.20\columnwidth}C{0.12\columnwidth}C{0.18\columnwidth}C{0.13\columnwidth}@{}}
\toprule
Term & GEE Wald $\chi^2$ (df) & GEE $p$ & Mixed $|z|>1.96$ & Median $|z|$ \\
\midrule
Model $\times$ contract & 214.58 (18) & $<10^{-4}$ & 18/18 & 10.56 \\
Contract $\times$ configuration & 29.25 (6) & $<10^{-4}$ & 6/6 & 6.19 \\
Model $\times$ configuration & 32.57 (12) & 0.0011 & 6/12 & 2.23 \\
\bottomrule
\end{tabularx}
}
\caption{Task-clustered logistic robustness. Rows are interaction terms; columns report GEE tests and mixed-model coefficient summaries.}
\label{tab:strict50-clustered-logistic}
\end{table}

Together, these tests support the performance profiles in \Cref{tab:model-runtime-matrix} and \Cref{fig:model-runtime-gap-heatmap}. The largest non-additive structure is model--contract specific, while the smaller exposure-configuration contribution remains sufficient to change some local configuration choices.

\section{Conclusion}

\benchmark{} measures the portability of reference scores, model selections, and exposure configurations across complete target contracts. Holding task semantics fixed, 12{,}600 matched episodes cover seven model-family representatives. Additive Proxy corrections leave substantial error and model-order reversals occur, while exposure transfer incurs smaller within-suite regret. These findings support validating reference-based decisions under target contracts.

\section{Limitations}

\benchmark{} is a fixed comparative instrument of 50 exactly checkable rule-conditioned decisions. Its primary role is controlled identification of reference-to-target decision validity under matched task semantics. Each evaluation setting is analyzed as a complete executable contract, with recorded outcomes supporting outcome reconstruction and provenance verification. The same matched design can extend to additional tasks, models, configurations, and contracts while preserving task, skill, seed, and checker alignment.

\clearpage
\bibliography{refs}

\clearpage
\appendix
\section{Measurement Suite and Contract Specifications}

This supplement details the controlled measurement suite, contract-indexed checked performance, and cross-contract conclusion portability. It provides task examples, contract and exposure specifications, uncertainty checks, and formal-analysis details. Upon acceptance, we will open-source the benchmark and evaluation code and publicly release the associated task data and run records.

\begin{table*}[!htbp]
\centering
{\small
\begin{tabularx}{\textwidth}{@{}P{0.17\textwidth}P{0.29\textwidth}P{0.10\textwidth}P{0.25\textwidth}Y@{}}
\toprule
Evaluation setting & Task categories & Tasks & Example tasks & Analysis role \\
\midrule
\benchmark{} & Memory (8), operations (8), GitHub (7), inbox (6), research (6), browser (5), calendar (5), state-dependent coordination (3), temporal state (1), and compliance (1). & 50 & Shared-inbox routing, durable memory repair, CI/release triage, multi-page browser collection, calendar reconciliation, deployment readiness, research table reconciliation, and security-questionnaire scoping. & Model--contract outcomes, exposure selection, and formal interaction analyses. \\
\bottomrule
\end{tabularx}
}
\caption{\benchmark{} composition: 50 controlled skill-dependent tasks covering ten recurring task categories. The same suite supports the outcome, exposure-configuration, and formal interaction analyses.}
\label{tab:inventory}
\end{table*}

\begin{table*}[!htbp]
\centering
{\small
\begin{tabularx}{\textwidth}{@{}P{0.12\textwidth}P{0.20\textwidth}P{0.22\textwidth}P{0.22\textwidth}Y@{}}
\toprule
Contract & Skill placement & Agent loop and state & Skill access interface & Recorded records \\
\midrule
\proxy{} & Selected skill cards are appended to the task prompt. & One chat-completion turn; no separate workspace state beyond the prompt. & Skill text is ordinary prompt context, so there is no separate workspace-access step. & Final answer, checker result, selected skill identities, run configuration, and token usage. \\
\official{} & Selected skills are installed as workspace skill files beside the task documents. & OpenClaw agent loop in an isolated workspace with file and harness tools enabled. & Skill content is reachable through the workspace interface. & Final answer, checker result, selected skill identities, default recorded harness configuration, run metadata, and token usage. \\
Hermes & The same selected skills are exposed through Hermes' external skill directory. & Hermes planning loop with file and skill tools and per-episode home/state. & The agent can list or view exposed skills through Hermes skill tools, or inspect workspace files. & Final answer, checker result, selected skill identities, default recorded harness configuration, and run metadata. \\
GenericAgent & The same selected skills are installed in the episode workspace. & GenericAgent loop with reset memory and file/tool actions under the same task workspace. & The agent can inspect files in the workspace; there is no separate skill-list API in the default recorded configuration. & Final answer, checker result, selected skill identities, default recorded harness configuration, and run metadata. \\
\bottomrule
\end{tabularx}
}
\caption{Execution semantics for the contracts compared in the paper. The matched protocol fixes the task, seed, deterministic selector result, selected skills, and checker; each target harness runs under its complete default recorded configuration. Skill placement, state, tools, access interface, and agent loop jointly define each recorded contract.}
\label{tab:runtime-semantics}
\end{table*}

Each matrix cell records its model, contract, seed, runtime configuration, timeout, and retry metadata. Proxy uses a 2{,}000-token output budget and recorded 300--900 second provider timeouts. OpenClaw runs \texttt{openclaw@2026.3.24}; Hermes and GenericAgent use 12-turn loops with recorded 180--900 second timeouts.

The offline construction audit accepts 50/50 reference answers and rejects 50/50 predefined rule-violating answers. Direct-answer leakage counts are zero across the prompt, filename, artifact-payload, and checker-visible channels.

\section{Composite Exposure-Configuration Protocol}

For each task, the suite defines a target skill identifier and a 16-item task-scoped candidate skill pool. Single Skill applies the deterministic scoped selector and exposes its highest-ranked candidate after merging equivalent identifiers. Shortlist follows the same selector ranking, removes unavailable and duplicate entries, and exposes up to four available candidate skills. Full Pool exposes all 16 task-scoped candidates. The selector is deterministic conditional on the task, seed, and seed-instantiated candidate context; the 150 selector decisions correspond to 50 tasks under three evaluation seeds.

Across 150 task--seed selector decisions, Single Skill contains the target skill in 135 cases and Shortlist in all 150. Across seven models and four settings, these correspond to 3{,}780/4{,}200 and 4{,}200/4{,}200 episode rows. The configurations compare top-1 commitment, a target-covering compact set, and complete task-scoped visibility.

\begin{table*}[!htbp]
\centering
{\small
\begin{tabular}{@{}lrrrrr@{}}
\toprule
Contract & Cells & Episodes & Clean episodes & Surfaced-ID records & Provenance files \\
\midrule
Proxy & 63 & 3{,}150 & 3{,}150 & 3{,}150 & 63 \\
OpenClaw & 63 & 3{,}150 & 3{,}150 & 3{,}150 & 63 \\
Hermes & 63 & 3{,}150 & 3{,}150 & 3{,}150 & 63 \\
GenericAgent & 63 & 3{,}150 & 3{,}150 & 3{,}150 & 63 \\
\bottomrule
\end{tabular}
}
\caption{Delivery-integrity audit for the broad matrix. All 3{,}150 matched model--configuration--seed--task groups have identical available-skill pools, selector retrieval records, and surfaced-skill identities across the four contracts, establishing the common delivery configuration used by the outcome analysis.}
\label{tab:delivery-integrity}
\end{table*}

\section{Natural-Language Descriptions of Benchmark Tasks}

The following natural-language descriptions summarize four benchmark tasks through their instruction, visible workspace evidence, selected reusable rule, expected answer value, and a plausible rule-violating alternative. The complete task instances, including their original files and structured answer specifications, will be released upon acceptance.

\subsection{GitHub: Public Release-Note Selection}

\paragraph{Instruction.} Choose the release-note bundle and return JSON only.

\paragraph{Visible workspace.} The merged items are a billing-export retry-copy update, a serializer cleanup for admin snapshot exports, and a limited-tenant rollout-default change for the next cohort. QA records that the retry copy changed in the export flow, the serializer difference appears on a debug surface, and the next cohort follows the new default. Support expects questions about the export wording; an admin team noticed the snapshot grouping; rollout operations asks whether the default change should be announced.

\paragraph{Selected reusable rule.} Public release notes include customer- or operator-visible changes, exclude changes restricted to debug/admin surfaces, and include a rollout-default change once it governs the next cohort.

\paragraph{Expected output.} \path{include_export_wording_change_and_limited_tenant_rollout_default_change}.

\paragraph{Plausible wrong output.} \path{include_export_wording_change_and_admin_snapshot_serializer_difference}.

\subsection{Compliance: Security-Questionnaire Scope}

\paragraph{Instruction.} Choose the forwardable questionnaire answer packet and return JSON only.

\paragraph{Visible workspace.} A customer asks whether Harbor Archive exports are covered by the current encryption-at-rest control. The evidence index contains a forwardable current control extract covering core Harbor services, an internal archive-export architecture note, a compensating-control attestation for the linked archive-export subservice, and a previously forwarded general answer. The service map distinguishes the archive-export subservice from core Harbor services.

\paragraph{Selected reusable rule.} A core-service control extract does not by itself cover the archive-export subservice, but a signed compensating-control attestation may support a scoped customer answer for a linked subservice.

\paragraph{Expected output.} \path{answer_yes_with_scoped_exception_note_and_compensating_control_attestation}.

\paragraph{Plausible wrong output.} \path{answer_yes_with_current_control_extract_only}.

\subsection{Research: Benchmark-Table Schema Reconciliation}

\paragraph{Instruction.} Choose the main-table action for a new benchmark run and return JSON only.

\paragraph{Visible workspace.} A new HarborBench run reports 74.2 under \texttt{macro\_by\_task}, while existing main-table runs report 72.8 and 73.1 under \texttt{macro\_by\_family}. The draft table makes direct replacement with the newest score tempting, while a comparison note leaves both appendix-only reporting and schema-normalized main-table update plausible.

\paragraph{Selected reusable rule.} A main-table row cannot be directly replaced by a score under a noncanonical schema. For this release, however, the benchmark defines a mapping from \texttt{macro\_by\_task} to the canonical \texttt{macro\_by\_family} reporting frame.

\paragraph{Expected output.} \path{apply_schema_mapping_then_update_main_table_with_74_2}.

\paragraph{Plausible wrong output.} \path{replace_main_table_row_directly_with_74_2}.

\subsection{Calendar: Invite Timezone Canonicalization}

\paragraph{Instruction.} Choose the outbound invite timezone action and return JSON only.

\paragraph{Visible workspace.} An external Wednesday review is already agreed, but the current draft renders it as 09:30 in the organizer's displayed pre-travel timezone. The partner accepts either host-basis or recipient-local labels, although mixed labels have caused confusion. The thread records travel-week timezone churn and leaves a basis rewrite under discussion.

\paragraph{Selected reusable rule.} After the travel boundary, outbound invites use the organizer's post-travel effective timezone; for this partner, the canonical label is the host basis rather than the recipient-local basis.

\paragraph{Expected output.} \path{rewrite_invite_into_post_travel_host_basis}.

\paragraph{Plausible wrong output.} \path{send_displayed_pretravel_basis_invite}.

\section{Calibration and Decision Portability}

The regret analysis relates cross-contract configuration agreement to the resulting target-contract success difference on the same 50-task suite. Regret is computed from the point estimates in \Cref{tab:broad-policy-regret}; task-bootstrap uncertainty for rank transfer and exposure selection is reported in \Cref{tab:broad-decision-bootstrap}.

\begin{table}[!htbp]
\centering
{\small
\setlength{\tabcolsep}{1.8pt}
\begin{tabular}{@{}lcccc@{}}
\toprule
Target contract & \makecell{Winner-set\\change} & \makecell{Positive\\regret} & Mean & Max \\
\midrule
OpenClaw & 4/7 & 3/7 & 0.048 & 0.167 \\
Hermes & 6/7 & 5/7 & 0.047 & 0.113 \\
GenericAgent & 7/7 & 4/7 & 0.017 & 0.073 \\
\bottomrule
\end{tabular}
}
\caption{Within-suite configuration agreement and regret across evaluation contracts. Regret is target-contract best success minus the target-contract success of the best configuration among the Proxy winner set on the same 50 tasks.}
\label{tab:broad-policy-regret}
\end{table}

The additive calibration analysis pairs each of the 189 target-contract cells with the Proxy cell for the same model, exposure configuration, and seed. The global model fits one offset in $\widehat{y}_{\mathrm{target}}=y_{\mathrm{Proxy}}+\delta$ across all paired cells; the contract-specific model fits one $\delta_h$ per target contract. The reported leave-one-model-out (LOMO) error estimates each offset from six model rows and evaluates it on the held-out row. The decision analysis resamples the 50 tasks 5{,}000 times while retaining all models, contracts, exposure configurations, and seeds for each sampled task. Rank intervals use tie-aware Kendall $\tau_b$ and count only strictly reversed pairs, so bootstrap ties are not broken by model identity.

\begin{table*}[!htbp]
\centering
{\small
\setlength{\tabcolsep}{1.6pt}
\begin{tabular}{@{}lccccc@{}}
\toprule
Method & Corr. & RMSE & MAE & Max & LOMO RMSE \\
\midrule
Uncalibrated Proxy & 0 & 0.496 & 0.444 & 0.920 & 0.496 \\
One global correction & 1 & 0.226 & 0.181 & 0.521 & 0.247 \\
One correction per contract & 3 & 0.188 & 0.159 & 0.506 & 0.214 \\
\bottomrule
\end{tabular}
}
\caption{Additive Proxy calibration over 189 matched target-contract cells. Corrections are fitted on target minus Proxy success. LOMO evaluates each model with corrections fitted on the other six model rows.}
\label{tab:broad-calibration}
\end{table*}

\begin{table*}[!htbp]
\centering
{\small
\setlength{\tabcolsep}{1.5pt}
\begin{tabular}{@{}lccccc@{}}
\toprule
& \multicolumn{2}{c}{Model-rank transfer} & \multicolumn{3}{c}{Exposure selection} \\
\cmidrule(lr){2-3} \cmidrule(lr){4-6}
Target & $\tau_b$ [95\%] & Rev. [95\%] & Changes [95\%] & Mean regret [95\%] & Positive \\
\midrule
OpenClaw & 0.238 [0.048, 0.488] & 8/21 [5, 10] & 4/7 [2, 6] & 0.048 [0.002, 0.088] & 3/7 \\
Hermes & 0.429 [0.150, 0.619] & 6/21 [4, 8] & 6/7 [3, 7] & 0.047 [0.013, 0.077] & 5/7 \\
GenericAgent & 0.333 [0.143, 0.524] & 7/21 [5, 9] & 7/7 [3, 7] & 0.017 [0.009, 0.050] & 4/7 \\
\bottomrule
\end{tabular}
}
\caption{Task-bootstrap stability of model-rank transfer and exposure selection (5{,}000 resamples). Rank intervals cover tie-aware Kendall's $\tau_b$ and strictly reversed-pair counts. Exposure intervals cover the number of changed Proxy-to-target winner sets and mean tie-robust regret across seven model rows.}
\label{tab:broad-decision-bootstrap}
\end{table*}

\section{Formal Analysis and Robustness}

The cell-level OLS response is cell success rate over \benchmark{}; cells are equally weighted because each aggregates the same 50 tasks. The additive model is $y_{m,r,p,s}=\alpha+\mu_m+\rho_r+\pi_p+\sigma_s+\epsilon_{m,r,p,s}$, where $m$, $r$, $p$, and $s$ index model, contract, exposure configuration, and seed. Each reported $\Delta R^2$ adds the named interaction block to the corresponding reduced model. The episode-level logistic model is $\mathrm{logit}\Pr(Y_i=1)=\alpha+\mu_{m_i}+\rho_{r_i}+\pi_{p_i}+\sigma_{s_i}+B_i$, where $B_i$ is the tested block. These checks are complemented by task-level bootstrap, leave-one-task-out analyses, GEE logistic models, and binomial mixed models with task random intercepts.

\begin{table*}[!htbp]
\centering
{\small
\setlength{\tabcolsep}{2.0pt}
\begin{tabular}{@{}lccc@{}}
\toprule
Added term & $\Delta R^2$ & Bootstrap & LOO \\
\midrule
All two-way interactions & 0.095 & 0.100 [0.083, 0.118] & 0.091--0.098 \\
Model $\times$ contract & 0.090 & 0.093 [0.076, 0.111] & 0.086--0.093 \\
Contract $\times$ configuration & 0.002 & 0.003 [0.001, 0.005] & 0.002--0.003 \\
Model $\times$ configuration & 0.003 & 0.004 [0.002, 0.008] & 0.003--0.004 \\
\bottomrule
\end{tabular}
}
\caption{Task-level resampling robustness for \benchmark{} interaction terms. Bootstrap resamples tasks; leave-one-task-out drops each task once.}
\label{tab:strict50-family-cluster-robustness}
\end{table*}

\end{document}